%% file: neurips_2025.tex
\documentclass{article}
\input{packages}

\usepackage[preprint]{neurips_2025}

\usepackage[utf8]{inputenc} 
\usepackage[T1]{fontenc}    
\usepackage{hyperref}       
\usepackage{url}            
\usepackage{booktabs}       
\usepackage{amsfonts}       
\usepackage{nicefrac}       
\usepackage{microtype}      
\usepackage{xcolor}         

\newcommand{\ours}{ETS\xspace}

\title{ETS: Efficient Tree Search for Inference-Time Scaling}

\author{%
  Coleman Hooper$^{1}$\enskip\enskip
  Sehoon Kim$^{1}$\enskip\enskip
  Suhong Moon$^{1}$\\[0.75mm]
  \textbf{
  Kerem Dilmen$^{1}$\enskip\enskip
  Monishwaran Maheswaran$^{1}$\enskip\enskip
  Nicholas Lee$^{1}$\enskip\enskip
  }\\[0.75mm]
  \textbf{
  Michael W. Mahoney$^{1,2,3}$\enskip\enskip
  Yakun Sophia Shao$^{1}$\enskip\enskip
  Kurt Keutzer$^{1}$\enskip\enskip
  Amir Gholami$^{1,2}$}\\[0.75mm]
  $^{1}$University of California, Berkeley \enspace
  $^{2}$ICSI \enspace
  $^{3}$LBNL \\[0.75mm]
  \texttt{\small{\{chooper, sehoonkim, suhong.moon, kerem.dilmen, monishwaran}}\\
  \texttt{\small{nicholas.lee, mahoneymw, ysshao, keutzer, amirgh\}@berkeley.edu}}\\
}

\begin{document}

\maketitle

\input{_0_abstract}
\input{_1_introduction}
\input{_2_related_work}
\input{_3_profiling}
\input{_4_algorithm}
\input{_5_results}
\input{_6_conclusion}

\bibliographystyle{plainnat}
\bibliography{main}

\input{_7_appendix}

\end{document}

%% file: packages.tex
\usepackage{soul}
\usepackage[utf8]{inputenc}
\usepackage{graphicx}
\usepackage{booktabs}

\usepackage{multicol}

\usepackage{graphicx}
\usepackage{booktabs} 

\newcommand\hd{ \rowcolor{teal!18}}

\usepackage[ruled,noend]{algorithm2e}

\SetCommentSty{mycommfont}

\usepackage{here}

\usepackage{amsmath,amssymb,amsfonts,amsbsy,amsfonts,latexsym}
\usepackage{multirow}
\usepackage{makecell}
\usepackage[labelfont=bf,textfont=it,belowskip=0pt,aboveskip=5pt,tableposition=top]{caption}
\usepackage{xcolor}
\usepackage{colortbl}

\usepackage{subcaption} 

\definecolor{colorA}{RGB}{189,201,225}
\definecolor{colorB}{RGB}{103,169,207}
\definecolor{colorC}{RGB}{ 28,144,153}
\definecolor{colorD}{RGB}{  1,108, 89}

\newcolumntype{R}{>{\columncolor{gray!40}}r}
\newcolumntype{L}{>{\columncolor{gray!40}}l}
\newcolumntype{C}{>{\columncolor{gray!40}}c}

\usepackage{tabularx,colortbl,xcolor}
\usepackage{multirow}
\usepackage[normalem]{ulem}
\useunder{\uline}{\ul}{}

\usepackage{enumitem}

\usepackage{xparse}

\DeclareGraphicsExtensions{.pdf,.png}

\SetKwInput{KwInput}{Input}

\usepackage{longtable}
\usepackage{pgfplots}
\usepackage{outlines}

%% file: _0_abstract.tex
\begin{abstract}
Test-time compute scaling has emerged as a new axis along which to improve model accuracy, where additional computation is used at inference time to allow the model to think longer for more challenging problems. 
One promising approach for test-time compute scaling is search against a process reward model, where a model generates multiple potential candidates at each step of the search, and these partial trajectories are then scored by a separate reward model in order to guide the search process.
The diversity of trajectories in the tree search process affects the accuracy of the search, since increasing diversity promotes more exploration.
However, this diversity comes at a cost, as divergent trajectories have less KV sharing, which means they consume more memory and slow down the search process.
Previous search methods either do not perform sufficient exploration, or else explore diverse trajectories but have high latency.
We address this challenge by proposing Efficient Tree Search (\ours), which promotes KV sharing by pruning redundant trajectories while maintaining necessary diverse trajectories. 
\ours incorporates a linear programming cost model to promote KV cache sharing by penalizing the number of nodes retained, while incorporating a semantic coverage term into the cost model to ensure that we retain trajectories which are semantically different.
We demonstrate how \ours can achieve \textbf{1.8}$\times$ reduction in average KV cache size during the search process, leading to \textbf{1.4}$\times$ increased throughput relative to prior state-of-the-art methods, with minimal accuracy degradation and without requiring any custom kernel implementation. 
Code is available at: \url{https://github.com/SqueezeAILab/ETS}. 

\end{abstract}

%% file: _1_introduction.tex
\begin{figure*}[h]
\centering
\includegraphics[width=\linewidth]{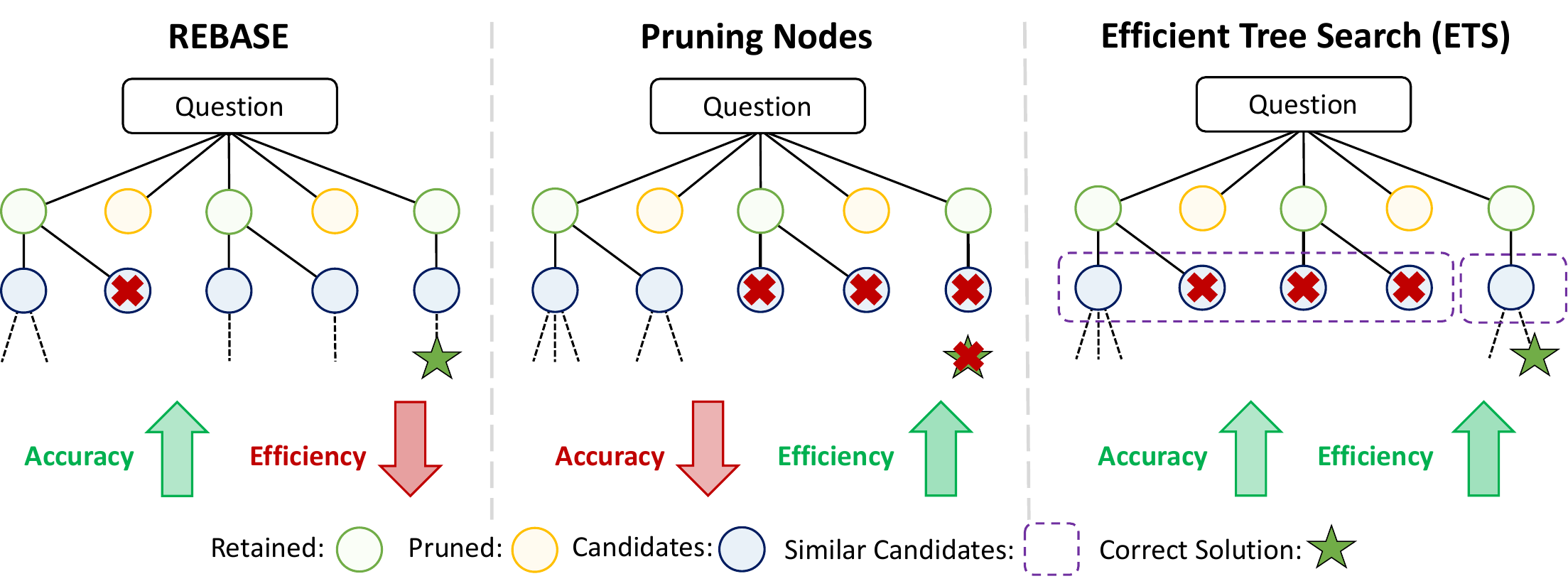}
 \caption{ 
  Visualization of our methodology for accelerating search while maintaining accuracy.
  (Left): Prior work (REBASE) achieves high accuracy with search by sampling more or less continuations from the candidate leaf nodes depending on their reward scores, thereby trading off exploration and exploitation. 
  However, this approach leads to larger KV cache size since it retains divergent trajectories which do not share nodes earlier in the tree.
  (Middle) One approach to reduce the cost is to penalize the number of nodes in the tree, thereby encouraging sharing prior KV cache state.
  However, this has the downside of pruning necessary semantically diverse trajectories, which degrades accuracy.
  (Right) \ours (our method) attains high efficiency by pruning out redundant nodes and promoting KV cache sharing, while ensuring that semantically diverse trajectories are retained, thereby allowing our approach to perform sufficient exploration.
  }
\vspace{-6mm}
  \label{fig:main}
\end{figure*}
\section{Introduction}

With the increasing challenges with scaling up language model pretraining due to compute and data availability, scaling test-time computation is emerging as a new axis along which to improve model performance \cite{snell2024scaling}. 
By increasing the amount of computation used at test time, we can allow Large Language Models (LLMs) to think longer for more challenging problems by generating long chain-of-thought reasoning, where the model lays out step-by-step reasoning to help guide itself to the final answer \cite{wei2022chain}.
This is particularly useful for improving accuracy in domains which require problem-solving capabilities, such as challenging math and coding problems \cite{snell2024scaling,brown2024large}.
One prominent method to scale up test time compute is through search, where we generate multiple possible candidate solutions and then filter them down to produce the final response.
An emerging method for performing search is to use a \textit{Process Reward Model} (PRM) to score partial trajectories in order to inform the search process \cite{snell2024scaling, beeching2024scalingtesttimecompute}.
These strategies generate multiple potential candidates at each step, score each of them using a PRM, and then choose which trajectories to retain based on these scores.
While these approaches yield promising results in terms of improving accuracy through test-time scaling, the increased inference costs are substantial, as many trajectories must be generated and scored before a final answer is selected.

Previous works on efficient search have either used number of model calls or FLOPs as the target efficiency metric when comparing search methods \cite{qiu2024treebon,wu2024inference, snell2024scaling}; however, these metrics may not be correlated with the actual inference costs of search.
In particular, prior work has highlighted that generative LLM inference is typically \textit{memory bandwidth-bound} \cite{kim2023squeezellm,kim2023full, hooper2024squeezed}, meaning that inference efficiency is limited by how quickly we can transfer data rather than the peak compute of the GPU (also referred to as memory wall~\cite{gholami2024ai}).
For tree search methods which sample multiple trajectories, each trajectory requires separate KV cache state, and the KV cache becomes a critical memory bottleneck as the width of the search increases.
A key challenge with proxy efficiency metrics such as model calls and FLOPs is that they neglect the influence of KV cache sharing between trajectories, which has a substantial impact on the memory requirements. 
For example, two trajectories which share the KV cache for most of their previous steps will require substantially fewer memory operations when performing further generations than if each trajectory has an entirely separate KV cache state.

In this work, we analyze the efficiency and performance of existing search methods, and identify a tradeoff between beam search and diverse tree search methods.
We focus in particular on parallel tree search algorithms like beam search and REBASE \cite{wu2024inference} since they are able to search multiple paths in parallel and are therefore more efficient for LLM test-time scaling than sequential search algorithms like MCTS \cite{snell2024scaling,beeching2024scalingtesttimecompute,wu2024inference}.
We find that beam search exhibits high KV cache sharing, but has low diversity and hence reduced accuracy even for wide beam widths.
Conversely, diverse sampling methods such as REBASE generate more diverse trajectories and are therefore able to attain higher accuracy and better scaling to wider beam widths.
However, these approaches lead to substantially reduced KV cache sharing, which in turn translates to substantial inference overheads.
In contrast with prior methods, our work achieves both high accuracy and high efficiency by proposing a search algorithm that promotes KV cache sharing to improve efficiency, while retaining the diverse trajectories that are necessary for facilitating exploration in order to attain high accuracy.
The main contributions of our work are as follows:

\begin{enumerate}[leftmargin=4mm]
    \item We present profiling data which demonstrates how existing proxy metrics for search efficiency are not necessarily correlated with search runtime due to the memory-bound nature of LLM inference as well as the impacts of KV cache sharing on attainable throughput.    
    \item 
    We design our algorithm to enhance the efficiency of tree search methods by encouraging KV cache sharing during the search process (visualized in Figure \ref{fig:main}).
    Our method incorporates a linear programming cost model to promote KV cache sharing by penalizing divergent branches, thereby improving inference efficiency. 
    \item We present Efficient Tree Search (\ours), our proposed search strategy which retains diversity while promoting KV sharing, as highlighted in Figure \ref{fig:main}. 
    \ours incorporates a cost model to promote KV cache sharing, and augments this cost model with a coverage term that compels the search process to retain semantically diverse trajectories.
    This approach allows our method to retain diverse trajectories while promoting KV cache sharing by pruning out semantically similar trajectories.
    \item We present benchmarking results demonstrating how \ours achieves \textbf{1.8}$\times$ reduction in average KV cache size during the search process, leading to \textbf{1.4}$\times$ speedups relative to REBASE with minimal accuracy degradation. Notably, our approach doesn't require dedicated kernel implementations on top of SGLang, and is compatible with the benefits attainable from more efficient kernels for computing attention with tree structured KV sharing \cite{yao2024deft}. 
\end{enumerate}

%% file: _2_related_work.tex
\section{Related Work}

\subsection{Rejection Sampling}
Previous works have explored methods where multiple independent trajectories are sampled for a particular problem, and we then select between them to determine the final answer \cite{wang2022self, chen2024more,beeching2024scalingtesttimecompute, brown2024large}.
These approaches are typically referred to as Best-of-N or rejection sampling. 
Once several responses have been generated by the model, we need to verify which of the final answers are correct.  
Prior work has proposed the use of a trained verifier to select between responses \cite{cobbe2021training}.
This verifier, or Outcome Reward Model (ORM), generates a reward score for each trajectory, and we can then select the best response based on this score.
Alternatively, we can leverage more complex aggregation strategies like majority voting or weighted majority voting (using the reward scores) to select the final response \cite{wang2022self, chen2024more,beeching2024scalingtesttimecompute}.
Prior work has also proposed training a verifier to decide whether each step in the solution process is correct, since this can provide finer granularity feedback in order to provide a better reward signal \cite{uesato2022solving}. 
These types of verifiers are called Process Reward Models (PRMs), and they have been shown to be more reliable than ORMs for domains such as math problem solving \cite{lightman2023let}.
One key challenge when designing a PRM is coming up with training data; prior work has aimed to automate the data generation process to facilitate training PRMs without large amounts of human labeled data \cite{wang2024math}.
Prior work has also leveraged a trained ORM to score partial responses in order to decompose the problem into steps \cite{light2025disc}.
Other alternative approaches instead use LLMs as reward models to guide the LLM post-training by providing feedback on the quality of generated outputs
\cite{bai2022constitutional,abdulhai2023lmrl,lee2024rlaif,pan2024autonomous}.

\subsection{Tree Search}

Tree-based search approaches have been  successful in applications such as games \cite{brown2017libratus, silver2016mastering, silver2017mastering}, and have also recently been extended to LLM reasoning to help navigate the space of possible trajectories \cite{yao2024tree,besta2024graph,zhou2023language}.
One approach for performing tree search with math problems is through the use of a PRM to  provide reward scores for partial trajectories, and to then use these scores to guide the search process \cite{snell2024scaling}.
There have also been several methods to improve the diversity and efficiency with search strategies.
One straightforward search strategy is beam search, which explores the solution space by sampling $N$ partial trajectories, pruning out all but $k$ of these trajectories based on the PRM scores, and then expanding each retained trajectory by sampling multiple continuations for each of them \cite{snell2024scaling,beeching2024scalingtesttimecompute, qiu2024treebon}.
This can attain higher accuracy for the same compute budget relative to Best-of-N sampling \cite{snell2024scaling}.
One challenge with naive beam search is that it doesn't generate diverse enough trajectories \cite{beeching2024scalingtesttimecompute}.
One proposed solution for this is Diverse Verifier Tree Search (DVTS), which first segments the beam search tree into multiple parallel subtrees, and then runs beam search independently within each subtree \cite{beeching2024scalingtesttimecompute}.
This is analogous to prior works on Diverse Beam Search for language model decoding \cite{li2016mutual,vijayakumar2016diverse}.
DVTS attains higher accuracy for wider beam widths than standard beam search; however, this has the potential additional cost of reduced KV sharing due to segmenting the tree into separate subtrees.

\subsection{Accelerating Tree Search}

There have been several prior approaches for accelerating tree search for LLM inference. 
REBASE \cite{wu2024inference} allocated different numbers of continuations for different partial trajectories based on the reward scores, thereby producing more diverse trajectories than standard beam search and attaining higher accuracy for the same efficiency budget (when measuring efficiency using FLOPs or number of model calls).
One alternative approach for accelerating search is to allocate different amounts of computation for different problems depending on their difficulty \cite{zhang2024scaling,snell2024scaling, beeching2024scalingtesttimecompute}.
Additionally, one prior work aiming to accelerate Best-of-N sampling terminated generations early using partial reward signals \cite{sun2024fast}.
There have also been several prior works that have designed support in serving systems for shared prefix or tree attention workloads.
Hydragen \cite{juravsky2024hydragen} proposed only storing the shared prefix for the KV cache once, and then batching together loads.
vLLM \cite{kwon2023efficient} also supports shared prefix workloads and avoids duplicating the KV cache.
SGLang \cite{zheng2023efficiently} supports Radix Attention, which compactly stores reused KV cache segments and dynamically references them, even for more complex tree sharing patterns.
Other works have also proposed efficient kernel implementations for computing attention with tree-structured KV sharing \cite{yao2024deft}.

%% file: _3_profiling.tex
\begin{figure*}[h]
\centering
\includegraphics[width=\linewidth]{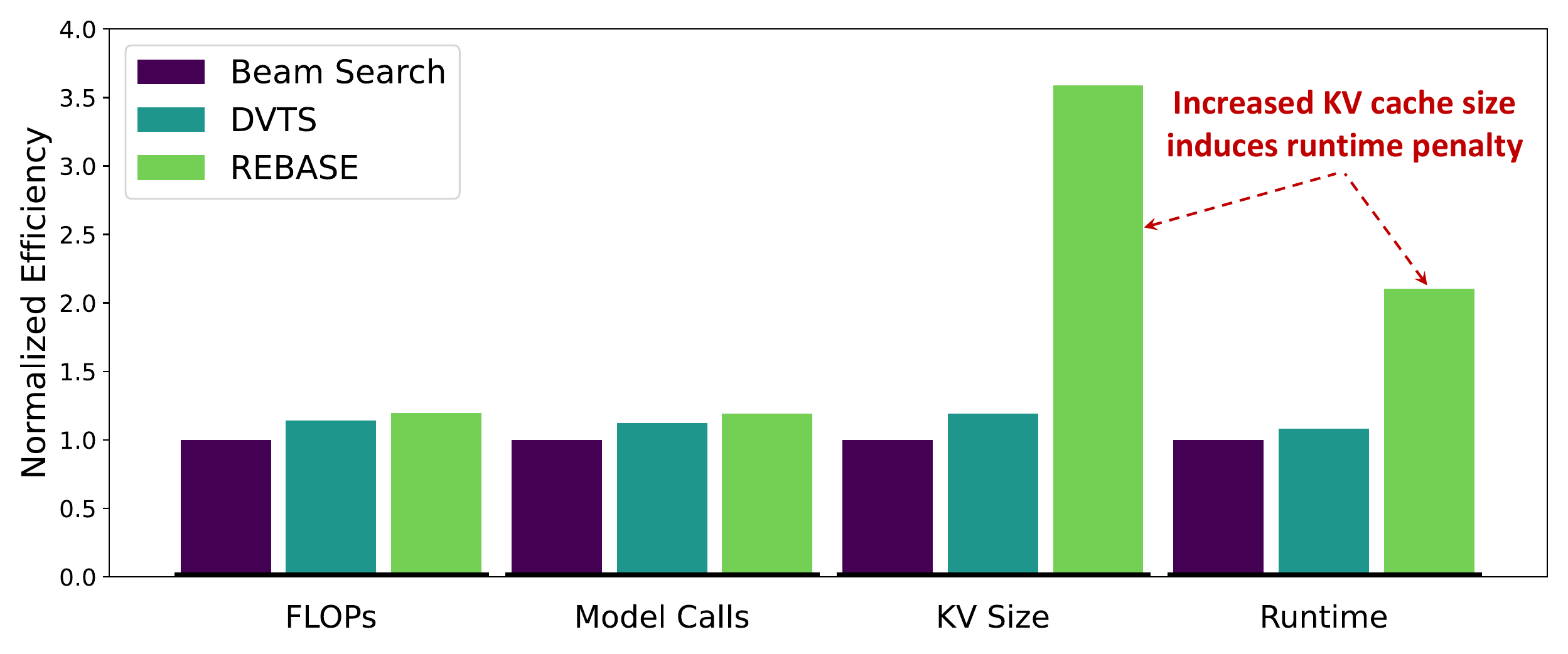}
\vspace*{-2mm}
 \caption{ 
 Correlation between approximate efficiency metrics and profiled runtime.
 We report FLOPs, number of model calls, and total KV Cache Size (``KV Size'') as well as profiled runtime.
 We measure each metric for Beam Search, DVTS, and REBASE for the Llemma-34B model with a width of 256, and we report each metric normalized to the value for Beam Search.
 For Beam Search and DVTS, we retain $\sqrt{N}$ trajectories at each step, where $N$ is the width of the search.
 As can be seen, REBASE has similar FLOPs and number of model calls compared to beam search and DVTS, but it exhibits significantly higher runtime.
 The increased runtime is due to its increased KV cache size.
 This clearly shows that FLOPs and number of model calls are not necessarily the right proxy metrics to use when assessing search efficiency.
  }
\vspace{-5mm}
  \label{fig:profiling}
\end{figure*}

\section{Profiling}

\label{sec:profiling}

When designing more efficient search algorithms, it is desirable to leverage a proxy metric to assess the efficiency of the algorithm in order to speed up iteration time, as running benchmarking in isolation for fair comparison between different search methods is more challenging and leads to slower iteration time.
Example proxy metrics for search runtime which were used in previous works include the number of model calls / tree width and the number of FLOPs \cite{qiu2024treebon,wu2024inference,snell2024scaling}.
The simplicity of these metrics makes them desirable for fast prototyping and ease of comparison.
However, as mentioned previously, FLOP count is not a good metric for memory-bound problems and may even be misleading.

To assess the utility of these proxy metrics, we profiled the throughput when running different search strategies, each using the same width.
We profiled throughput on 100 samples from the MATH500 test set on NVIDIA H100 NVL GPUs, with the Llemma-34B model and Llemma-Reward-34B PRM each on a separate GPU (model details are provided in Section \ref{sec:experimental-details}).
Results were collected by running evaluation using 8 parallel threads (which is analogous to the attainable throughput with a batch size of 8 for serving use-cases).
We also include FLOPs, model calls, and KV cache size as proxy metrics (where KV cache size is the sum of the size of the KV cache state across all steps in the search process).
To estimate FLOPs, we leverage the approximation that the number of FLOPs is proportional to the number of tokens generated (which holds for short context lengths \cite{pope2023efficiently}).
According to existing proxy metrics, these search methods should all yield similar latency as they all have similar numbers of FLOPs and model calls.
However, as highlighted in Figure \ref{fig:profiling}, we observe that these algorithms exhibit substantial differences in terms of profiled latency.
These gaps are due to the differences in terms of number of \textit{memory operations} which need to be performed when running the workload, since generative LLM inference is typically memory bandwidth-bound \cite{kim2023full,kim2023squeezellm, hooper2024squeezed}.
The number of memory operations that need to be performed for different tree search methods is dependent on multiple factors: 

\begin{enumerate}[leftmargin=4mm]
    \item The amount of KV cache sharing between trajectories determines the size of the state that needs to be loaded for each inference step, which substantially influences the runtime if the KV cache is the dominant contributor to memory consumption relative to model weights. 
    Note that leveraging this benefit requires efficient tree attention kernels to reduce KV cache loads \cite{yao2024deft}; however, these kernels are not yet integrated within efficient serving frameworks like SGLang.
    \item 
    Even with existing open-source frameworks, such as SGLang~\cite{zheng2023efficiently}, that don't incorporate tree attention kernels, we can still reduce memory operations.
    If the KV cache for the sequences is too large to fit in memory (either with wide beam search or in batched use-cases), then the number of sequences that can be run in parallel will be constrained and the search process gets fragmented into multiple successive iterations. 
    This leads to performing more memory operations for the model weights, since the model weights need to be loaded for each fragment.
    \item Additionally, while the KV cache state can be reused for earlier steps in the search when generating later steps, if the memory requirements are too great then the KV cache for the earlier steps would be de-allocated and would need to be recomputed, which would increase latency.
\end{enumerate}

These factors mean that reducing the KV cache size by promoting KV cache sharing between trajectories is crucial for reducing latency, even though the KV cache size is \textit{not} considered by existing proxy metrics.
As such, \textbf{we argue that KV cache size serves as a superior proxy metric for assessing efficiency of different strategies (assuming equal search width)}.

%% file: _4_algorithm.tex
\section{Algorithm}

\subsection{Encouraging KV Sharing}
\label{sec:algo-kv}

As shown in Section \ref{sec:profiling}, the efficiency of search strategies is strongly influenced by the amount of available KV cache sharing among the retained trajectories. 
However, the most accurate search strategy, REBASE \cite{wu2024inference}, has substantial inference overheads due to the reduction in KV cache sharing from sampling in a more balanced manner when deciding which trajectories to expand.
In order to enhance the available KV cache sharing, we design a cost model-based method which we integrate at each step in the search process when determining which trajectories to sample continuations from.
This cost model is designed to mitigate the efficiency overheads from balanced sampling (which is necessary for obtaining diverse trajectories) by adding a cost term which accounts for the overhead of divergent KV branches in the search tree. 
The objective of our algorithm is outlined in Figure \ref{fig:main} (Middle), where we aim to promote KV cache sharing by minimizing the number of retained nodes in order to minimize the cost of the search process.

A key challenge with incorporating efficiency considerations when selecting which trajectories to retain is that we cannot consider each trajectory in isolation. 
For example, if trajectories $A$ and $B$ share node $C$ in the search tree (and therefore share the corresponding KV cache state), we need to retain node $C$ if we retain \textit{either} trajectory $A$ or trajectory $B$.
Due to the importance of KV sharing in determining runtime, this means that when deciding whether to keep trajectory $A$, the efficiency implications of pruning trajectory $A$ are only apparent if we determine which other trajectories are retained or pruned. 
The efficiency cost of retaining each trajectory therefore depends \textit{on the other trajectories that are retained or pruned}.
As such, it is challenging to assign a cost for retaining each trajectory, and we can only account for efficiency \textit{at the scope of the tree}.

To account for efficiency in the search process, our algorithm therefore decides which trajectories to retain or prune while considering the efficiency costs and benefits of retaining a set of trajectories, 
rather than assigning an efficiency cost and reward for each separate trajectory.
In order to facilitate this, we formulate our objective as an integer linear programming (ILP) problem, 
where we have a binary variable corresponding to whether each trajectory is retained.
We include additional constraint binary variables for each earlier node in the search tree.
For each node, we set the corresponding binary variable to 1 if any of the trajectories that share this node are retained.
This allows us to optimize the ILP to consider interdependency between trajectories when determining the amount of exploitable KV sharing.

Our cost model needs to contain both a term which factors in the reward for keeping each trajectory, as well as the cost of the set of trajectories that we select.
For the reward term in our cost model, we use the weights obtained with REBASE~\cite{wu2024inference} sampling as the value for retaining each trajectory.
As in the open-source implementation for \cite{wu2024inference}, the weights are determined by iterating over the trajectories from highest-reward to lowest-reward, and computing the weight $W_i$ as follows, where $N_i$ is the  number of continuations that still need to be sampled, $R_i$ is the reward for trajectory $i$ computed using a Process Reward Model (PRM), $T_R$ is a temperature parameter that controls how balanced the sampling is, and $A_i$ is the set of all remaining trajectories for which we have not yet sampled continuations:

\vspace{-0.5\baselineskip}
\begin{equation}
    \label{eq:rebase}
    W_i = \mathrm{ceil} ( N_i \frac{\mathrm{exp}(R_i / T_R)}{\sum_{k \in A_i}\mathrm{exp}(R_k / T_R)}  )
\end{equation}
\vspace{-0.5\baselineskip}

This sampling procedure produces more continuations for more promising trajectories (as determined by their reward scores), while still producing some continuations from less promising trajectories (as opposed to beam search, which retains a subset of promising trajectories and then prunes the rest).
By still producing some continuations for less promising trajectories, REBASE performs additional exploration, which yields higher accuracy for the same search width.

We then introduce the following objective which we want to maximize at each step, which contains one term which retains the highest priority trajectories, as well as a second term that promotes KV sharing by penalizing the size of the retained tree:

\vspace{-0.5\baselineskip}
\begin{equation}
    \label{eq:cost_model_v1}
    \max_{S}( \frac{\sum_{i \in S} W_i}{\sum_{i \in A} W_i} 
   \;-\;
   \lambda_b \,\frac{|V_S|}{|V_A|} )
\end{equation}
\vspace{-0.5\baselineskip}

where $S$ is the set of selected trajectories (which can be variable in size);
$V_A$ and $V_S$ are, respectively, the set of all nodes in the tree before applying the cost model and the set of nodes that are retained after pruning the tree; and $\lambda_b$ is a hyperparameter which controls the relative strength of the budget term. 
Note that we add the additional constraint that $|S| \geq 1$ to ensure that we always retain at least one leaf node.
This objective aims to maximize the reward of the retained nodes, while still maintaining efficiency.
Also, note that if $\lambda_b$ is set to 0, this objective bears similarity to the optimization target for existing strategies like beam search, which aim to maximize the reward for the retained set of trajectories (under some constraint for the number of trajectories that can be kept).
For $\lambda_b > 0$, this term penalizes the number of retained nodes in the tree, which encourages retaining trajectories which share part of their prior KV cache state rather than completely distinct trajectories.
As outlined in Section \ref{sec:appendix-ilp-kv}, this problem can be formulated as a constrained ILP by using binary decision variables indicating whether each node in the tree is retained or not.
To solve this ILP, we leverage the Pulp optimization library \cite{mitchell2011pulp} using the CBC solver \cite{john_forrest_2024_13347261}.

After we apply our KV cache-promoting cost model to prune out trajectories, we then re-apply REBASE sampling to determine how many continuations to sample from each of the remaining trajectories.
The updated weights for the subsequent iteration are then set by iterating over the selected trajectories from highest-reward to lowest-reward and computing the updated weight $W'_i$ as follows, where $S_i$ is the set of retained trajectories for which we have not yet sampled continuations:

\vspace{-1\baselineskip}
\begin{equation}
    \label{eq:rebase_v2}
    W'_i = \mathrm{ceil} ( N_i \frac{\mathrm{exp}(R_i / T_R)}{\sum_{k \in S_i}\mathrm{exp}(R_k / T_R)}  )
\end{equation}
\vspace{-1\baselineskip}

\begin{figure*}[h]
\centering
\includegraphics[width=\linewidth]{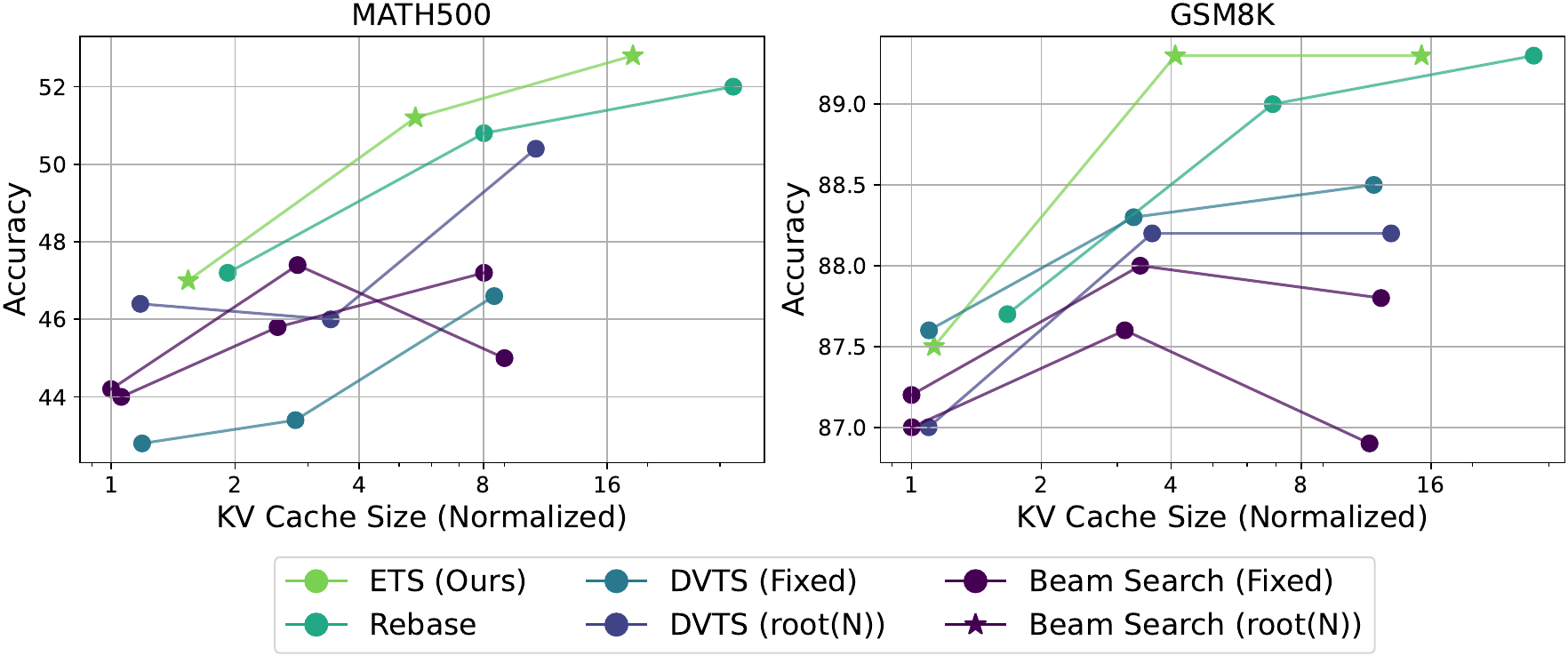}
 \caption{ 
 Accuracy versus efficiency trade-off curves for different search strategies with the Llemma-34B model.
 We report results for search widths of 16, 64, and 256 across all methods.
 We provide baseline results for Beam Search and DVTS (both with retaining a fixed number of trajectories as well as $\sqrt{N}$ trajectories at each step) \cite{snell2024scaling,beeching2024scalingtesttimecompute}, as well as for REBASE \cite{wu2024inference}.
 Our results demonstrate how our method allows for improved efficiency relative to REBASE, while maintaining the accuracy benefits due to retaining necessary diverse trajectories.
  }
  \label{fig:results}
\vspace{-5mm}
\end{figure*}
 
\subsection{Retaining Diversity}

\label{sec:algo-cluster}

The crucial challenge as we encourage KV cache sharing is the importance of diversity for attaining high accuracy.
If we naively enforce KV sharing, this will also consequently prune out diverse trajectories which were crucial for attaining high accuracy.
This creates a difficult trade-off, where we want to retain diverse trajectories to maintain accuracy, but not at the expense of large runtime penalties.

However, existing approaches for sampling continuations lead to many redundant or similar continuations.
Even if some of these continuations are not exact duplicates, they can still have a similar semantic meaning.
For example, consider the first steps of two different solution trajectories for the question: ``The results of a cross-country team's training run are graphed below. Which student has the greatest average speed?"

\vspace{-0.5\baselineskip}
\begin{enumerate}
    \item \textit{Step 1: The average speed is the total distance divided by the total time.}
    \item \textit{Step 1: To find the average speed, we need to divide the distance traveled by the time taken.}
\end{enumerate}
\vspace{-0.5\baselineskip}

Although these two steps are phrased differently, they convey the exact same meaning. This redundancy suggests that retaining both is unnecessary.
Several of these continuations could be pruned out without losing out on diversity.
Our goal is therefore to retain trajectories that are semantically diverse,
while improving KV cache sharing by removing redundant or similar trajectories which are unnecessary for attaining high accuracy. 

To ensure that we retain diverse trajectories, we augment our cost model approach from Section \ref{sec:algo-kv} with an additional term that promotes retaining diverse trajectories.
The goal is for the set of selected trajectories to cover the original semantic space of the sampled trajectories.
However, similar to the challenges with estimating the impact of retaining a given trajectory on exploitable KV sharing, we can only estimate whether retaining a given trajectory improves coverage when considering the other trajectories that are retained.
One possible approach for considering the diversity of trajectories that are retained when pruning to reduce KV cache size would be to estimate the pairwise similarity between trajectories and to then add a term to the cost model that penalizes pairwise similarity among the set of retained trajectories. 
However, this approach would additionally introduce quadratic terms in the cost model, making the optimization objective quadratic rather than linear.
Incorporating pairwise similarity terms would also make it infeasible to solve the cost model for large beam widths.

To more efficiently estimate the coverage of our selected trajectories without incorporating pairwise similarity computation directly into the cost model, we first cluster the trajectories and then estimate diversity of retained trajectories based on whether we retain trajectories across different clusters. 
Our approach is visualized in Figure \ref{fig:main} (Right), where we ensure that semantically diverse trajectories are retained when we apply our pruning method to reduce the size of the tree.
This approach ensures that we retain \textit{coverage} over the original semantic space of potential trajectories.
To accomplish this, we first embed the last step for each sequence using a BERT model finetuned for embedding math sentences \cite{steinfeldt2024evaluation}.
We then cluster these embeddings using  hierarchical agglomerative clustering from Scipy \cite{virtanen2020scipy} based on cosine similarity.
We use a fixed distance threshold in order to determine the number of clusters.
Note that our choice of similarity metric is arbitrary, and our algorithm is also compatible with alternate methods for measuring semantic similarity between the trajectories.

After clustering trajectories, we then incorporate this into our objective as a coverage term, which accounts for how well our selected set of trajectories covers the possible solution space.
We incorporate the diversity term into the objective function which we want to maximize at each step as follows:

\vspace{-0.5\baselineskip}
\begin{equation}
    \label{eq:cost_model_v2}
    \max_S( \frac{\sum_{i \in S} W_i}{\sum_{i \in A} W_i} 
   \;-\;
   \lambda_b \,\frac{|V_S|}{|V_A|} + \lambda_d \frac{|C_S|}{|C_A|})
\end{equation}
\vspace{-0.5\baselineskip}

where $C_S$ and $C_A$ refer to the clusters covered by the selected trajectories and the total number of clusters, respectively; and $\lambda_d$ is a hyperparameter which controls the relative strength of the coverage term.
This term ensures that we retain sufficient semantic diversity in order to facilitate exploration during the search process, which is critical for achieving high accuracy.
Section \ref{sec:appendix-ilp-cluster}, describes how this can be incorporated into our previously formulated ILP such that we can solve this optimization problem efficiently.
As in Section \ref{sec:algo-kv}, after we apply our cost model to prune out trajectories, we re-apply REBASE \cite{wu2024inference} to sample continuations from each of the remaining trajectories. 

%% file: _5_results.tex
\begin{table*}[t]
\centering
\caption{
Accuracy versus KV cache size for REBASE as well as \ours. 
Results are provided for MATH500 and GSM8K for the Llemma-34B, Mistral-7B-SFT, and Llama-3.2-1B-Instruct models.
We report the KV cache size reduction (``KV Red.'') for each width (relative to REBASE), where higher is better since it implies a reduction in memory consumption.
}
\label{tab:results}
\scriptsize
\setlength{\tabcolsep}{5pt}
\renewcommand{\arraystretch}{1.2}
%
\begin{subtable}{0.47\linewidth}
    \centering
    \begin{tabular}{ccccccc}
    \toprule
    \multirow{2}{*}{\textbf{Method}} & \multicolumn{2}{c}{\textbf{Width=16}} & \multicolumn{2}{c}{\textbf{Width=64}} & \multicolumn{2}{c}{\textbf{Width=256}} \\
    \cline{2-7}
     & \textbf{Acc.} & \textbf{KV $\downarrow$} & \textbf{Acc.} & \textbf{KV $\downarrow$} & \textbf{Acc.} & \textbf{KV $\downarrow$} \\
    \midrule
    \multicolumn{7}{c}{\textbf{Llemma-34B}} \\
    \midrule
    REBASE       &   47.2         & 1$\times$   &           50.8 & 1$\times$  &   52.0         & 1$\times$  \\
    \hd \ours   &  47.0          & 1.2$\times$  & 51.2          & 1.5$\times$  &    52.8        & 1.8$\times$  \\

    \midrule
    \multicolumn{7}{c}{\textbf{Mistral-7B-SFT}} \\
    \midrule
    REBASE      &     38.8       & 1$\times$   &  43.4          & 1$\times$   &     42.4       & 1$\times$   \\
    \hd \ours   &   39.4         & 1.3$\times$  &  43.2          & 1.3$\times$  &    42.2        & 1.7$\times$  \\
    \midrule
    \multicolumn{7}{c}{\textbf{Llama-3.2-1B-Instruct}} \\
    \midrule
    REBASE      &     48.0       & 1$\times$   &  52.6          & 1$\times$   &     53.8       & 1$\times$   \\
    \hd \ours   &   48.0         & 1.2$\times$  &  52.4          & 1.5$\times$  &    53.8        & 1.6$\times$  \\
    \bottomrule
    \end{tabular}
    \newline
    \newline
    \small\textbf{MATH500}
\end{subtable}
\hspace{3mm}
\begin{subtable}[t]{0.47\linewidth}
    \centering
    \begin{tabular}{ccccccc}
    \toprule
    \multirow{2}{*}{\textbf{Method}} & \multicolumn{2}{c}{\textbf{Width=16}} & \multicolumn{2}{c}{\textbf{Width=64}} & \multicolumn{2}{c}{\textbf{Width=256}} \\
    \cline{2-7}
     & \textbf{Acc.} & \textbf{KV $\downarrow$} & \textbf{Acc.} & \textbf{KV $\downarrow$} & \textbf{Acc.} & \textbf{KV $\downarrow$} \\
    \midrule
    \multicolumn{7}{c}{\textbf{Llemma-34B}} \\
    \midrule
    REBASE       &    87.7        & 1$\times$   &   89.0         & 1$\times$   &      89.3      & 1$\times$   \\
    \hd \ours   &    87.5        & 1.5$\times$  &   89.3         & 1.7$\times$ &   89.3          & 1.8$\times$  \\
    \midrule
    \multicolumn{7}{c}{\textbf{Mistral-7B-SFT}} \\
    \midrule
    REBASE       &     88.6       & 1$\times$   &  89.1          & 1$\times$   &   90.1        & 1$\times$   \\
    \hd \ours   &    88.3        & 1.2$\times$  &   89.2         &  1.6$\times$  &     89.6       & 1.3$\times$  \\
    \midrule
    \multicolumn{7}{c}{\textbf{Llama-3.2-1B-Instruct}} \\
    \midrule
    REBASE       &     82.3      & 1$\times$   &  85.9          & 1$\times$   &   88.1        & 1$\times$   \\
    \hd \ours   &    82.7        & 1.1$\times$  &   85.8         &  1.2$\times$  &     88.3       & 1.3$\times$  \\
    \bottomrule
    \end{tabular}
    \newline
    \newline
    \small\textbf{GSM8K}
\end{subtable}
 \vspace{-6mm}
\end{table*}

\section{Evaluation}

\subsection{Experimental Details}

\label{sec:experimental-details}

We leverage the open-source REBASE code for the balanced sampling implementation \cite{wu2024inference}, and we use SGLang for serving models \cite{zheng2023efficiently}.
For all experiments in our evaluation, we leverage temperature sampling with a temperature of 1.0 and we fix the REBASE temperature at 0.2, which are the default settings in the open-source code for \cite{wu2024inference}.
We use the final PRM score at each step as the reward for that step, and we select the final answer with weighted majority voting using the final PRM score for each trajectory as the weight.
We use these aggregation strategies since they have been shown to outperform other methods of aggregating trajectories to determine the final response \cite{beeching2024scalingtesttimecompute}.
As in \cite{wu2024inference}, we reduce the search width each time a retained trajectory completes.
We set $\lambda_d = 1$ throughout our evaluation, and we sweep over $\lambda_b \in [1,2]$ (with increasing $\lambda_b$ corresponding to more aggressive KV compression) and select the largest value of $\lambda_b$ which doesn't degrade accuracy by greater than $0.2\%$.

We provide results for three groups of models. We leverage the Llemma-34B model (finetuned on Metamath) along with the Llemma-34B PRM from \cite{wu2024inference}.
We  use the Mistral-7B model finetuned on Metamath as well as the corresponding Mistral-7B PRM from \cite{wang2024math}.
Finally, we leverage the Llama-3.2-1B-Instruct model \cite{grattafiori2024llama} along with the Llama3.1-8B-PRM-Deepseek-Data PRM \cite{dong2024rlhf}.
We report results on MATH500 and GSM8K \cite{cobbe2021training} with search widths of 16, 64, and 256.

We compare against several baseline search strategies.
We include results for beam search both with 4 trajectories retained at each step and with  $\sqrt{N}$ trajectories retained at each step, where $N$ is the initial width of the search, as in \cite{snell2024scaling}.
We also include comparisons with DVTS both with 4 trajectories retained at each step and with $\sqrt{N}$ trajectories retained at each step (where the number of trajectories retained at each step is also the same as the number of separate subtrees), as in \cite{beeching2024scalingtesttimecompute}.
Finally, we provide comparisons against REBASE, which serves as our strongest baseline due to its high accuracy relative to search width \cite{wu2024inference}.

\subsection {Results}

Figure \ref{fig:results} shows the accuracy results relative to efficiency for \ours (in terms of total KV cache size). We provide results for the Llemma-34B model for both MATH500 and GSM8K datasets.
We also report results for the beam search, DVTS, and REBASE baseline methods.
Our results demonstrate that our approach, which considers both diversity and efficiency, is able to attain a better accuracy versus efficiency trade-off than existing search strategies. 
Table \ref{tab:results} also shows results on both MATH500 as well as GSM8K for the Llemma-34B, Mistral-7B, and Llama-3.2-1B-Instruct models.
We provide accuracy results as well as profiled KV cache compression estimates.
These results highlight how the benefits of our search strategy are consistent across different model families and datasets, as we are able to maintain accuracy while obtaining consistent efficiency benefits relative to the REBASE baseline \cite{wu2024inference}.

\subsection{Throughput Benchmarking}

\begin{table}[t!]
\caption{
Throughput for our approach relative to REBASE \cite{wu2024inference}, measured using 100 samples from the MATH500 test set with the Llemma-34B model on H100 GPUs (with the beam width set to 256).
We report throughput improvements using a beam width of 256.
We also include the reduction in KV cache size (normalized to REBASE), as well as the accuracy for each approach.
}
\scriptsize
\label{tab:throughput}
\centering{
\begin{tabular}{cccc}
\toprule
   \textbf{Method} & \textbf{Accuracy} & \textbf{KV $\downarrow$} & \textbf{Throughput}\\
   \midrule
   \midrule
   REBASE  & 52.0 & 1$\times$ & 1$\times$  \\
   
   \hd \textbf{\ours}  & \textbf{52.8} & \textbf{1.8}$\times$ &  \textbf{1.4}$\times$ \\
\bottomrule
\end{tabular}
}
\vspace{-5mm}
\end{table}

Table \ref{tab:throughput} provides measured throughput for REBASE as well as \ours on H100 NVL GPUs.
We benchmark both REBASE and \ours using [4,8,16,32] parallel threads (which is representative for the serving scenario with a batch size equal to the number of threads) and select the best configuration for each.
We run benchmarking with the main LLM and the PRM each on a separate H100 NVL GPU, and for \ours we co-locate the embedding model on the same GPU as the reward model.
We observe {1.4}$\times$ increased throughput relative to the baseline REBASE method, demonstrating how the increased KV cache sharing from our algorithm translates to higher throughput, without requiring any custom kernels.
Appendix \ref{sec:appendix-rt} also reports the low overhead of running the ILP solver and the embedding model for clustering, which consumes under 2\% of overall runtime.

%% file: _6_conclusion.tex
\section{Conclusion}

Computational efficiency is a key bottleneck for exploiting test-time scaling in order to enhance model accuracy.
An emerging approach for exploiting test-time scaling is through tree search against a verifier.
A key challenge with existing tree search methods is the trade-off between efficiency and accuracy; high accuracy with tree search necessitates diverse trajectories, but retaining diverse trajectories leads to high inference costs due to reduced KV cache sharing in the tree.
We perform profiling which demonstrates the importance of KV cache sharing, and show that existing efficiency metrics like FLOPs and model calls are insufficient for assessing the efficiency trade-offs between tree search methods due to the impacts of KV sharing.
We then propose \ours, a search strategy which promotes KV cache sharing while retaining diverse trajectories in order to attain high accuracy.
\ours encourages KV cache sharing in the tree search by penalizing divergent branches in the tree.
Our method also incorporates a coverage term which ensures that semantically diverse trajectories are maintained while pruning redundant trajectories, 
thereby allowing for sufficient exploration to retain the accuracy benefits of diverse tree search.
The combination of these components of our method allows us to retain necessary diversity while pruning out redundancy in order to enable accurate and efficient tree search.
\ours achieves \textbf{1.8}$\times$ reduction in average KV cache size during the search process, which translates to \textbf{1.4}$\times$ increased throughput, with minimal accuracy loss and without requiring custom kernel implementations.
Our method demonstrates the potential of leveraging efficiency considerations during search to enable accurate and efficient search for test-time scaling.

\section{Acknowledgements}
We acknowledge gracious support from the FuriosaAI team including Jihoon Yoon, Suyeol Lee, and Hyung Il Koo, as well as from Intel, Apple, NVIDIA, and Mozilla.
We also appreciate the support from Microsoft through their Accelerating Foundation Model Research, including great support from Sean Kuno.
Furthermore, we appreciate support from
Google Cloud, the Google TRC team, and specifically Jonathan Caton, and Prof. David Patterson.
Prof. Keutzer's lab is sponsored by the Intel corporation, UC Berkeley oneAPI Center of Excellence, Intel VLAB team, as well as funding through BDD and BAIR.
We appreciate great feedback and support from Ellick Chan, Saurabh Tangri, Andres Rodriguez, and Kittur Ganesh.
Sehoon Kim and Suhong Moon would like to acknowledge the support from the Korea Foundation for Advanced Studies (KFAS).
Michael W. Mahoney would also like to acknowledge
a J. P. Morgan Chase Faculty Research Award as well as the DOE, NSF, and ONR.
This work was supported by the Director, Office of Science, Office of Advanced Scientific Computing Research, of the U.S. Department of Energy under Contract No. DE-AC02-05CH11231.
Our conclusions do not necessarily reflect the position or the policy of our sponsors, and no official endorsement should be~inferred.

%% file: _7_appendix.tex
\newpage
\appendix

\section{Limitations}
\label{sec:appendix-limitations}

One limitation of our work is that the approach requires the use of a trained process reward model (PRM).
This means that our method depends on the generalization capabilities of the PRM in order for our approach to be used for different downstream tasks.
One potential avenue for extending our work is to leverage instruction-tuned models to obtain a reward signal to guide the search process, without using a trained PRM, thereby allowing our method to be used flexibly for different downstream tasks.
Another limitation is that the evaluation results in our work are constrained to mathematical reasoning tasks.
Important future work would be extending our memory-efficient search algorithm to other downstream problem-solving tasks like code generation.

\section{ILP Formulation}
\label{sec:appendix-ilp}

In this section, we first outline the setup for the optimization target in Section \ref{sec:algo-kv} to reduce KV cache memory usage, and then outline how we refine this target in Section \ref{sec:algo-cluster} to ensure that we retain necessary semantically diverse trajectories.
We formulate this problem as an ILP which we solve to determine which nodes to prune and which nodes to sample continuations from at each step in the search. 

\subsection{ILP Formulation for Encouraging KV Cache Sharing}
\label{sec:appendix-ilp-kv}

The initial optimization target described in Equation \ref{eq:cost_model_v1} encourages KV cache sharing by incorporating two terms: one term which is the sum of the retained weights post-pruning (to preferentially keep the most promising nodes), and another term which is the total number of retained nodes in the tree (to minimize the cost of the search). 
We use the weights obtained using REBASE~\cite{wu2024inference} sampling as the value for retaining each trajectory.
As in the open-source implementation for \cite{wu2024inference}, the weights are determined by iterating over the trajectories from highest-reward to lowest-reward, and computing the weight $W_i$ as follows, where $N_i$ is the  number of continuations that still need to be sampled, $R_i$ is the reward for trajectory $i$ computed using a Process Reward Model (PRM), $T_R$ is a temperature parameter that controls how balanced the sampling is, and $A_i$ is the set of all remaining trajectories for which we have not yet sampled continuations:

\begin{equation}
    \label{eq:rebase}
    W_i = \mathrm{ceil} ( N_i \frac{\mathrm{exp}(R_i / T_R)}{\sum_{k \in A_i}\mathrm{exp}(R_k / T_R)}  )
\end{equation}

To formulate our optimization problem as an ILP, we first build a list of all unique nodes in the tree, and create a binary variable for each node in the tree ($x$ for each of the leaf nodes, and $y$ for each of the non-leaf nodes) which represents whether we keep that node. 
We associate the weight for each leaf node with the corresponding binary variable $x$ for that leaf node.
For each non-leaf node, we identify all of its descendant leaf nodes (which are the latest steps in the search for any trajectory that shares this node), and we then constrain $y$ for this non-leaf node to be 1 if $x$ is 1 for any of its descendant leaf nodes, and we set $y$ to 0 otherwise. 
This implementation ensures that for each non-leaf node, $y$ is set to 1 if any of the descendant leaf nodes are retained (since we would then need to retain this node), whereas if all descendant leaf nodes are pruned then $y$ is set to 0 (as this node can also be pruned).
We then solve the ILP by setting the binary variables $x$ and $y$ to maximize the sum of the weights for the retained leaf nodes, while minimizing the portion of $x$  and $y$ that are set to 1 to minimize cost.

Let $L$ be the number of leaf nodes and $P$ be the number of non-leaf nodes. Given the binary variables $x_i \in \{0, 1\},  \forall i \in \{1, \dots, L\}$ and 
$y_j \in \{0, 1\},  \forall j \in \{1, \dots, P\}$, we formulate the initial ILP as follows:

\begin{equation}
    \label{eq:ilp_v1}
    \max_{x_i, y_j}
    \underbrace{\frac{\sum_{i=1}^{L} W_i x_i}{\sum_{i=1}^{L} W_i}}_{\text{Retained Weights}}
    \;-\; 
    \lambda_b \cdot
    \underbrace{\frac{\sum_{j = 1}^{P} y_j + \sum_{i = 1}^{L} x_i}{P+L}}_{\text{Budget Penalty}}
\end{equation}

with the following constraint that we keep non-leaf nodes if any of their descendent leaf nodes are retained:
\begin{align*}
&\text{Keep non-leaf node: } && y_j \geq x_i && \forall j \in \{1, \dots, P\}, \forall i \in \text{descendant}(j) \\
&\text{Keep at least one leaf node: } && \sum_{i=1}^{L} x_i \geq 1 &&  \\
\end{align*}
\subsection{ILP Formulation for Retaining Semantically Diverse Trajectories}
\label{sec:appendix-ilp-cluster}

In order to ensure that we retain necessary diverse trajectories while promoting KV cache sharing, we incorporate an additional term into our optimization target (as shown in Equation \ref{eq:cost_model_v2}), which represents coverage across semantic clusters. 
To compute this term, we first embed the last step for each trajectory and cluster the final steps based on cosine similarity.
To incorporate this term into our ILP formulation from Appendix \ref{sec:appendix-ilp-kv}, we add a binary variable $z$ for each cluster, which we constrain to be 1 if $x$ is 1 for any of the leaf nodes in this cluster, and 0 otherwise.
We then incorporate an additional term in the optimization target which represents the portion of semantic clusters which are covered by the set of retained trajectories. 

Let $L$ be the number of leaf nodes and $P$ be the number of non-leaf nodes, and let $K$ be the number of clusters.
Given the binary variables $x_i \in \{0, 1\}, \forall i \in \{1, \dots, L\}$, $y_j \in \{0, 1\}, \forall j \in  \{1, \dots, P\}$, and $z_k \in \{0, 1\}, \forall k \in \{1, \dots, K\}$, we formulate the updated ILP as follows:

\begin{equation}
    \label{eq:ilp_v2}
    \max_{x_i, y_j, z_k} 
    \underbrace{\frac{\sum_{i=1}^{L} W_i x_i}{\sum_{i=1}^{L} W_i}}_{\text{Retained Weights}}
    \;-\; 
    \lambda_b \cdot
    \underbrace{\frac{\sum_{j = 1}^{P} y_j + \sum_{i = 1}^{L} x_i}{P+L}}_{\text{Budget Penalty}}
    \;+\;
    \lambda_d \cdot
    \underbrace{\frac{1}{K}\sum_{k=1}^{K} z_k}_{\text{Semantic Coverage}} 
\end{equation}

with the following constraints:
\begin{align*}
&\text{Keep non-leaf node: } && y_j \geq x_i && \forall j \in \{1, \dots, P\}, \forall i \in \text{descendant}(j) \\
&\text{Cluster coverage:} && z_k \leq x_i && \forall k \in \{1, \dots, K\}, \forall i \in C_k \\
&\text{Keep at least one leaf node: } && \sum_{i=1}^{L} x_i \geq 1 &&  \\
\end{align*}

All together, our method encourages KV cache sharing by penalizing the cost of the tree (i.e. the number of nodes retained), while preferentially retaining semantically diverse trajectories in order to maintain coverage of the semantic space of sampled solutions.

\section{Clustering and ILP Overhead}
\label{sec:appendix-rt}

To assess the overhead of using a BERT model for embedding each trajectory, as well as performing clustering and running the ILP, we profiled the portion of runtime spent in these components
versus the amount of time spent running generation and the reward model.
We leverage the same experimental setup as described in Section \ref{sec:experimental-details}, and measured results for the Llemma-34B model using a search width of 256.
We profiled the overhead of these components using a batch size of 1, and averaged the time across all steps in the search process.
We found that the overhead of the embedding model, clustering, and running the ILP is only 2\% of the total runtime. 
This demonstrates the low overhead of the embedding model, clustering method, and ILP solver. 
We also replicated this analysis for the batched serving context (with 32 threads) to ensure there were no overheads from contention on CPU resources (or on the embedding model).
We found that the overhead in this case was also less than 2\% of the total runtime (from the perspective of each thread). 
Additionally, the runtime overheads can be hidden in the batched inference context, as the next step of the search can be running for one sample in the batch while we compute embeddings and run clustering for another sample.

%% file: neurips_2025.bbl
\begin{thebibliography}{42}
\providecommand{\natexlab}[1]{#1}
\providecommand{\url}[1]{\texttt{#1}}
\expandafter\ifx\csname urlstyle\endcsname\relax
  \providecommand{\doi}[1]{doi: #1}\else
  \providecommand{\doi}{doi: \begingroup \urlstyle{rm}\Url}\fi

\bibitem[Abdulhai et~al.(2023)Abdulhai, White, Snell, Sun, Hong, Zhai, Xu, and Levine]{abdulhai2023lmrl}
Marwa Abdulhai, Isadora White, Charlie Snell, Charles Sun, Joey Hong, Yuexiang Zhai, Kelvin Xu, and Sergey Levine.
\newblock Lmrl gym: Benchmarks for multi-turn reinforcement learning with language models.
\newblock \emph{arXiv preprint arXiv:2311.18232}, 2023.

\bibitem[Bai et~al.(2022)Bai, Kadavath, Kundu, Askell, Kernion, Jones, Chen, Goldie, Mirhoseini, McKinnon, et~al.]{bai2022constitutional}
Yuntao Bai, Saurav Kadavath, Sandipan Kundu, Amanda Askell, Jackson Kernion, Andy Jones, Anna Chen, Anna Goldie, Azalia Mirhoseini, Cameron McKinnon, et~al.
\newblock Constitutional ai: Harmlessness from ai feedback.
\newblock \emph{arXiv preprint arXiv:2212.08073}, 2022.

\bibitem[Beeching et~al.(2024)Beeching, Tunstall, and Rush]{beeching2024scalingtesttimecompute}
Edward Beeching, Lewis Tunstall, and Sasha Rush.
\newblock Scaling test-time compute with open models, 2024.
\newblock URL \url{https://huggingface.co/spaces/HuggingFaceH4/blogpost-scaling-test-time-compute}.

\bibitem[Besta et~al.(2024)Besta, Blach, Kubicek, Gerstenberger, Podstawski, Gianinazzi, Gajda, Lehmann, Niewiadomski, Nyczyk, et~al.]{besta2024graph}
Maciej Besta, Nils Blach, Ales Kubicek, Robert Gerstenberger, Michal Podstawski, Lukas Gianinazzi, Joanna Gajda, Tomasz Lehmann, Hubert Niewiadomski, Piotr Nyczyk, et~al.
\newblock Graph of thoughts: Solving elaborate problems with large language models.
\newblock In \emph{Proceedings of the AAAI Conference on Artificial Intelligence}, volume~38, pages 17682--17690, 2024.

\bibitem[Brown et~al.(2024)Brown, Juravsky, Ehrlich, Clark, Le, R{\'e}, and Mirhoseini]{brown2024large}
Bradley Brown, Jordan Juravsky, Ryan Ehrlich, Ronald Clark, Quoc~V Le, Christopher R{\'e}, and Azalia Mirhoseini.
\newblock Large language monkeys: Scaling inference compute with repeated sampling.
\newblock \emph{arXiv preprint arXiv:2407.21787}, 2024.

\bibitem[Brown et~al.(2017)Brown, Sandholm, and Machine]{brown2017libratus}
Noam Brown, Tuomas Sandholm, and Strategic Machine.
\newblock Libratus: The superhuman ai for no-limit poker.
\newblock In \emph{IJCAI}, pages 5226--5228, 2017.

\bibitem[Chen et~al.(2024)Chen, Davis, Hanin, Bailis, Stoica, Zaharia, and Zou]{chen2024more}
Lingjiao Chen, Jared~Quincy Davis, Boris Hanin, Peter Bailis, Ion Stoica, Matei Zaharia, and James Zou.
\newblock Are more llm calls all you need? towards the scaling properties of compound ai systems.
\newblock In \emph{The Thirty-eighth Annual Conference on Neural Information Processing Systems}, 2024.

\bibitem[Cobbe et~al.(2021)Cobbe, Kosaraju, Bavarian, Chen, Jun, Kaiser, Plappert, Tworek, Hilton, Nakano, et~al.]{cobbe2021training}
Karl Cobbe, Vineet Kosaraju, Mohammad Bavarian, Mark Chen, Heewoo Jun, Lukasz Kaiser, Matthias Plappert, Jerry Tworek, Jacob Hilton, Reiichiro Nakano, et~al.
\newblock Training verifiers to solve math word problems.
\newblock \emph{arXiv preprint arXiv:2110.14168}, 2021.

\bibitem[Dong et~al.(2024)Dong, Xiong, Pang, Wang, Zhao, Zhou, Jiang, Sahoo, Xiong, and Zhang]{dong2024rlhf}
Hanze Dong, Wei Xiong, Bo~Pang, Haoxiang Wang, Han Zhao, Yingbo Zhou, Nan Jiang, Doyen Sahoo, Caiming Xiong, and Tong Zhang.
\newblock Rlhf workflow: From reward modeling to online rlhf.
\newblock \emph{arXiv preprint arXiv:2405.07863}, 2024.

\bibitem[Forrest et~al.(2024)Forrest, Ralphs, Vigerske, Santos, Forrest, Hafer, Kristjansson, jpfasano, EdwinStraver, Jan-Willem, Lubin, rlougee, a~andre, jpgoncal1, Brito, h-i gassmann, Cristina, Saltzman, tosttost, Pitrus, MATSUSHIMA, Vossler, SWGY, and to~st]{john_forrest_2024_13347261}
John Forrest, Ted Ralphs, Stefan Vigerske, Haroldo~Gambini Santos, John Forrest, Lou Hafer, Bjarni Kristjansson, jpfasano, EdwinStraver, Jan-Willem, Miles Lubin, rlougee, a~andre, jpgoncal1, Samuel Brito, h-i gassmann, Cristina, Matthew Saltzman, tosttost, Bruno Pitrus, Fumiaki MATSUSHIMA, Patrick Vossler, Ron~@ SWGY, and to~st.
\newblock coin-or/cbc: Release releases/2.10.12, August 2024.
\newblock URL \url{https://doi.org/10.5281/zenodo.13347261}.

\bibitem[Gholami et~al.(2024)Gholami, Yao, Kim, Hooper, Mahoney, and Keutzer]{gholami2024ai}
Amir Gholami, Zhewei Yao, Sehoon Kim, Coleman Hooper, Michael~W Mahoney, and Kurt Keutzer.
\newblock Ai and memory wall.
\newblock \emph{IEEE Micro}, 2024.

\bibitem[Grattafiori et~al.(2024)Grattafiori, Dubey, Jauhri, Pandey, Kadian, Al-Dahle, Letman, Mathur, Schelten, Vaughan, et~al.]{grattafiori2024llama}
Aaron Grattafiori, Abhimanyu Dubey, Abhinav Jauhri, Abhinav Pandey, Abhishek Kadian, Ahmad Al-Dahle, Aiesha Letman, Akhil Mathur, Alan Schelten, Alex Vaughan, et~al.
\newblock The llama 3 herd of models.
\newblock \emph{arXiv preprint arXiv:2407.21783}, 2024.

\bibitem[Hooper et~al.(2024)Hooper, Kim, Mohammadzadeh, Maheswaran, Paik, Mahoney, Keutzer, and Gholami]{hooper2024squeezed}
Coleman Hooper, Sehoon Kim, Hiva Mohammadzadeh, Monishwaran Maheswaran, June Paik, Michael~W Mahoney, Kurt Keutzer, and Amir Gholami.
\newblock Squeezed attention: Accelerating long context length llm inference.
\newblock \emph{arXiv preprint arXiv:2411.09688}, 2024.

\bibitem[Juravsky et~al.(2024)Juravsky, Brown, Ehrlich, Fu, R{\'e}, and Mirhoseini]{juravsky2024hydragen}
Jordan Juravsky, Bradley Brown, Ryan Ehrlich, Daniel~Y Fu, Christopher R{\'e}, and Azalia Mirhoseini.
\newblock Hydragen: High-throughput llm inference with shared prefixes.
\newblock \emph{arXiv preprint arXiv:2402.05099}, 2024.

\bibitem[Kim et~al.(2023{\natexlab{a}})Kim, Hooper, Gholami, Dong, Li, Shen, Mahoney, and Keutzer]{kim2023squeezellm}
Sehoon Kim, Coleman Hooper, Amir Gholami, Zhen Dong, Xiuyu Li, Sheng Shen, Michael~W Mahoney, and Kurt Keutzer.
\newblock Squeezellm: Dense-and-sparse quantization.
\newblock \emph{arXiv preprint arXiv:2306.07629}, 2023{\natexlab{a}}.

\bibitem[Kim et~al.(2023{\natexlab{b}})Kim, Hooper, Wattanawong, Kang, Yan, Genc, Dinh, Huang, Keutzer, Mahoney, et~al.]{kim2023full}
Sehoon Kim, Coleman Hooper, Thanakul Wattanawong, Minwoo Kang, Ruohan Yan, Hasan Genc, Grace Dinh, Qijing Huang, Kurt Keutzer, Michael~W Mahoney, et~al.
\newblock Full stack optimization of transformer inference: a survey.
\newblock \emph{arXiv preprint arXiv:2302.14017}, 2023{\natexlab{b}}.

\bibitem[Kwon et~al.(2023)Kwon, Li, Zhuang, Sheng, Zheng, Yu, Gonzalez, Zhang, and Stoica]{kwon2023efficient}
Woosuk Kwon, Zhuohan Li, Siyuan Zhuang, Ying Sheng, Lianmin Zheng, Cody~Hao Yu, Joseph Gonzalez, Hao Zhang, and Ion Stoica.
\newblock Efficient memory management for large language model serving with pagedattention.
\newblock In \emph{Proceedings of the 29th Symposium on Operating Systems Principles}, pages 611--626, 2023.

\bibitem[Lee et~al.(2024)Lee, Phatale, Mansoor, Lu, Mesnard, Ferret, Bishop, Hall, Carbune, and Rastogi]{lee2024rlaif}
Harrison Lee, Samrat Phatale, Hassan Mansoor, Kellie~Ren Lu, Thomas Mesnard, Johan Ferret, Colton Bishop, Ethan Hall, Victor Carbune, and Abhinav Rastogi.
\newblock {RLAIF}: Scaling reinforcement learning from human feedback with {AI} feedback, 2024.
\newblock URL \url{https://openreview.net/forum?id=AAxIs3D2ZZ}.

\bibitem[Li and Jurafsky(2016)]{li2016mutual}
Jiwei Li and Dan Jurafsky.
\newblock Mutual information and diverse decoding improve neural machine translation.
\newblock \emph{arXiv preprint arXiv:1601.00372}, 2016.

\bibitem[Light et~al.(2025)Light, Cheng, Yue, Oyamada, Wang, Paternain, and Chen]{light2025disc}
Jonathan Light, Wei Cheng, Wu~Yue, Masafumi Oyamada, Mengdi Wang, Santiago Paternain, and Haifeng Chen.
\newblock Disc: Dynamic decomposition improves llm inference scaling.
\newblock \emph{arXiv preprint arXiv:2502.16706}, 2025.

\bibitem[Lightman et~al.(2023)Lightman, Kosaraju, Burda, Edwards, Baker, Lee, Leike, Schulman, Sutskever, and Cobbe]{lightman2023let}
Hunter Lightman, Vineet Kosaraju, Yura Burda, Harri Edwards, Bowen Baker, Teddy Lee, Jan Leike, John Schulman, Ilya Sutskever, and Karl Cobbe.
\newblock Let's verify step by step.
\newblock \emph{arXiv preprint arXiv:2305.20050}, 2023.

\bibitem[Mitchell et~al.(2011)Mitchell, OSullivan, and Dunning]{mitchell2011pulp}
Stuart Mitchell, Michael OSullivan, and Iain Dunning.
\newblock Pulp: a linear programming toolkit for python.
\newblock \emph{The University of Auckland, Auckland, New Zealand}, 65:\penalty0 25, 2011.

\bibitem[Pan et~al.(2024)Pan, Zhang, Tomlin, Zhou, Levine, and Suhr]{pan2024autonomous}
Jiayi Pan, Yichi Zhang, Nicholas Tomlin, Yifei Zhou, Sergey Levine, and Alane Suhr.
\newblock Autonomous evaluation and refinement of digital agents.
\newblock \emph{arXiv preprint arXiv:2404.06474}, 2024.

\bibitem[Pope et~al.(2023)Pope, Douglas, Chowdhery, Devlin, Bradbury, Heek, Xiao, Agrawal, and Dean]{pope2023efficiently}
Reiner Pope, Sholto Douglas, Aakanksha Chowdhery, Jacob Devlin, James Bradbury, Jonathan Heek, Kefan Xiao, Shivani Agrawal, and Jeff Dean.
\newblock Efficiently scaling transformer inference.
\newblock \emph{Proceedings of Machine Learning and Systems}, 5:\penalty0 606--624, 2023.

\bibitem[Qiu et~al.(2024)Qiu, Lu, Zeng, Guo, Geng, Wang, Huang, Wu, and Wang]{qiu2024treebon}
Jiahao Qiu, Yifu Lu, Yifan Zeng, Jiacheng Guo, Jiayi Geng, Huazheng Wang, Kaixuan Huang, Yue Wu, and Mengdi Wang.
\newblock Treebon: Enhancing inference-time alignment with speculative tree-search and best-of-n sampling.
\newblock \emph{arXiv preprint arXiv:2410.16033}, 2024.

\bibitem[Silver et~al.(2016)Silver, Huang, Maddison, Guez, Sifre, Van Den~Driessche, Schrittwieser, Antonoglou, Panneershelvam, Lanctot, et~al.]{silver2016mastering}
David Silver, Aja Huang, Chris~J Maddison, Arthur Guez, Laurent Sifre, George Van Den~Driessche, Julian Schrittwieser, Ioannis Antonoglou, Veda Panneershelvam, Marc Lanctot, et~al.
\newblock Mastering the game of go with deep neural networks and tree search.
\newblock \emph{nature}, 529\penalty0 (7587):\penalty0 484--489, 2016.

\bibitem[Silver et~al.(2017)Silver, Hubert, Schrittwieser, Antonoglou, Lai, Guez, Lanctot, Sifre, Kumaran, Graepel, et~al.]{silver2017mastering}
David Silver, Thomas Hubert, Julian Schrittwieser, Ioannis Antonoglou, Matthew Lai, Arthur Guez, Marc Lanctot, Laurent Sifre, Dharshan Kumaran, Thore Graepel, et~al.
\newblock Mastering chess and shogi by self-play with a general reinforcement learning algorithm.
\newblock \emph{arXiv preprint arXiv:1712.01815}, 2017.

\bibitem[Snell et~al.(2024)Snell, Lee, Xu, and Kumar]{snell2024scaling}
Charlie Snell, Jaehoon Lee, Kelvin Xu, and Aviral Kumar.
\newblock Scaling llm test-time compute optimally can be more effective than scaling model parameters.
\newblock \emph{arXiv preprint arXiv:2408.03314}, 2024.

\bibitem[Steinfeldt and Mihaljevi{\'c}(2024)]{steinfeldt2024evaluation}
Christian Steinfeldt and Helena Mihaljevi{\'c}.
\newblock Evaluation and domain adaptation of similarity models for short mathematical texts.
\newblock In \emph{International Conference on Intelligent Computer Mathematics}, pages 241--260. Springer, 2024.

\bibitem[Sun et~al.(2024)Sun, Haider, Zhang, Yang, Qiu, Yin, Wang, Bartlett, and Zanette]{sun2024fast}
Hanshi Sun, Momin Haider, Ruiqi Zhang, Huitao Yang, Jiahao Qiu, Ming Yin, Mengdi Wang, Peter Bartlett, and Andrea Zanette.
\newblock Fast best-of-n decoding via speculative rejection.
\newblock \emph{arXiv preprint arXiv:2410.20290}, 2024.

\bibitem[Uesato et~al.(2022)Uesato, Kushman, Kumar, Song, Siegel, Wang, Creswell, Irving, and Higgins]{uesato2022solving}
Jonathan Uesato, Nate Kushman, Ramana Kumar, Francis Song, Noah Siegel, Lisa Wang, Antonia Creswell, Geoffrey Irving, and Irina Higgins.
\newblock Solving math word problems with process-and outcome-based feedback.
\newblock \emph{arXiv preprint arXiv:2211.14275}, 2022.

\bibitem[Vijayakumar et~al.(2016)Vijayakumar, Cogswell, Selvaraju, Sun, Lee, Crandall, and Batra]{vijayakumar2016diverse}
Ashwin~K Vijayakumar, Michael Cogswell, Ramprasath~R Selvaraju, Qing Sun, Stefan Lee, David Crandall, and Dhruv Batra.
\newblock Diverse beam search: Decoding diverse solutions from neural sequence models.
\newblock \emph{arXiv preprint arXiv:1610.02424}, 2016.

\bibitem[Virtanen et~al.(2020)Virtanen, Gommers, Oliphant, Haberland, Reddy, Cournapeau, Burovski, Peterson, Weckesser, Bright, et~al.]{virtanen2020scipy}
Pauli Virtanen, Ralf Gommers, Travis~E Oliphant, Matt Haberland, Tyler Reddy, David Cournapeau, Evgeni Burovski, Pearu Peterson, Warren Weckesser, Jonathan Bright, et~al.
\newblock Scipy 1.0: fundamental algorithms for scientific computing in python.
\newblock \emph{Nature methods}, 17\penalty0 (3):\penalty0 261--272, 2020.

\bibitem[Wang et~al.(2024)Wang, Li, Shao, Xu, Dai, Li, Chen, Wu, and Sui]{wang2024math}
Peiyi Wang, Lei Li, Zhihong Shao, Runxin Xu, Damai Dai, Yifei Li, Deli Chen, Yu~Wu, and Zhifang Sui.
\newblock Math-shepherd: Verify and reinforce llms step-by-step without human annotations.
\newblock In \emph{Proceedings of the 62nd Annual Meeting of the Association for Computational Linguistics (Volume 1: Long Papers)}, pages 9426--9439, 2024.

\bibitem[Wang et~al.(2022)Wang, Wei, Schuurmans, Le, Chi, Narang, Chowdhery, and Zhou]{wang2022self}
Xuezhi Wang, Jason Wei, Dale Schuurmans, Quoc Le, Ed~Chi, Sharan Narang, Aakanksha Chowdhery, and Denny Zhou.
\newblock Self-consistency improves chain of thought reasoning in language models.
\newblock \emph{arXiv preprint arXiv:2203.11171}, 2022.

\bibitem[Wei et~al.(2022)Wei, Wang, Schuurmans, Bosma, Xia, Chi, Le, Zhou, et~al.]{wei2022chain}
Jason Wei, Xuezhi Wang, Dale Schuurmans, Maarten Bosma, Fei Xia, Ed~Chi, Quoc~V Le, Denny Zhou, et~al.
\newblock Chain-of-thought prompting elicits reasoning in large language models.
\newblock \emph{Advances in neural information processing systems}, 35:\penalty0 24824--24837, 2022.

\bibitem[Wu et~al.(2024)Wu, Sun, Li, Welleck, and Yang]{wu2024inference}
Yangzhen Wu, Zhiqing Sun, Shanda Li, Sean Welleck, and Yiming Yang.
\newblock Inference scaling laws: An empirical analysis of compute-optimal inference for llm problem-solving.
\newblock In \emph{The 4th Workshop on Mathematical Reasoning and AI at NeurIPS'24}, 2024.

\bibitem[Yao et~al.(2024{\natexlab{a}})Yao, Chen, Zhang, You, Yuan, Wang, and Lin]{yao2024deft}
Jinwei Yao, Kaiqi Chen, Kexun Zhang, Jiaxuan You, Binhang Yuan, Zeke Wang, and Tao Lin.
\newblock Deft: Flash tree-attention with io-awareness for efficient tree-search-based llm inference.
\newblock \emph{arXiv preprint arXiv:2404.00242}, 2024{\natexlab{a}}.

\bibitem[Yao et~al.(2024{\natexlab{b}})Yao, Yu, Zhao, Shafran, Griffiths, Cao, and Narasimhan]{yao2024tree}
Shunyu Yao, Dian Yu, Jeffrey Zhao, Izhak Shafran, Tom Griffiths, Yuan Cao, and Karthik Narasimhan.
\newblock Tree of thoughts: Deliberate problem solving with large language models.
\newblock \emph{Advances in Neural Information Processing Systems}, 36, 2024{\natexlab{b}}.

\bibitem[Zhang et~al.(2024)Zhang, Zhou, Wang, Wang, and Li]{zhang2024scaling}
Kexun Zhang, Shang Zhou, Danqing Wang, William~Yang Wang, and Lei Li.
\newblock Scaling llm inference with optimized sample compute allocation.
\newblock \emph{arXiv preprint arXiv:2410.22480}, 2024.

\bibitem[Zheng et~al.(2023)Zheng, Yin, Xie, Huang, Sun, Yu, Cao, Kozyrakis, Stoica, Gonzalez, et~al.]{zheng2023efficiently}
Lianmin Zheng, Liangsheng Yin, Zhiqiang Xie, Jeff Huang, Chuyue Sun, Cody\_Hao Yu, Shiyi Cao, Christos Kozyrakis, Ion Stoica, Joseph~E Gonzalez, et~al.
\newblock Efficiently programming large language models using sglang.
\newblock 2023.

\bibitem[Zhou et~al.(2023)Zhou, Yan, Shlapentokh-Rothman, Wang, and Wang]{zhou2023language}
Andy Zhou, Kai Yan, Michal Shlapentokh-Rothman, Haohan Wang, and Yu-Xiong Wang.
\newblock Language agent tree search unifies reasoning acting and planning in language models.
\newblock \emph{arXiv preprint arXiv:2310.04406}, 2023.

\end{thebibliography}
